\documentclass{article}
\usepackage[utf8]{inputenc}
\usepackage[T1]{fontenc}
\usepackage{times} 
\usepackage{amsmath, amssymb} 
\usepackage{graphicx}
\usepackage[margin=1in]{geometry}
\usepackage{natbib} 
\usepackage{booktabs, tabularx}
\usepackage{hyperref}
\usepackage{microtype} 
\usepackage{algorithm, algorithmic}
\usepackage{enumitem}
\usepackage{fancyhdr}
\usepackage{float}
\usepackage{setspace}
\usepackage{ulem} 

\newcommand{\keywords}[1]{\textbf{Keywords:} #1}
\title{Optimizing Multi-Tier Supply Chain Ordering with a Hybrid Liquid Neural Network and Extreme Gradient Boosting Model}
\author{Chunan Tong, CSCP-F \\ University of Maryland, College Park\\ College Park, MD, USA \\ Email: tcn1989@umd.edu} 

\begin{document}
\onehalfspacing

\maketitle
\begin{abstract}
Supply chain management (SCM) encounters major obstacles, such as demand variability, inventory mismatches, and increased upstream order fluctuations caused by the bullwhip effect. Conventional approaches, like simple moving averages, find it challenging to cope with ever-changing market conditions. The Vending Machine Test is a key benchmark for LLMs, simulating real-world vending machine sales prediction scenarios, but Grok-4 and many other AI models still struggle with the complex continuous time series data in SCM, making it hard to accurately predict vending machine sales that fluctuate hourly, daily or seasonally. However, new machine learning (ML) methods, including LSTM, reinforcement learning, and XGBoost, present possible solutions but are hindered by computational complexity, training inefficiencies, or limitations in time-series modeling. Liquid Neural Networks (LNN), drawing inspiration from dynamic biological systems, offer a promising alternative due to their adaptability, low computational demands, and resilience to noise, making them ideal for real-time decision-making and edge computing. Although they have been successful in areas like autonomous vehicles and medical monitoring, their potential in optimizing supply chains is still largely untapped. This study introduces a hybrid LNN+XGBoost model to refine ordering strategies in multi-tier supply chains. By utilizing LNN’s dynamic feature extraction and XGBoost’s global optimization strengths, the model seeks to reduce the bullwhip effect and boost overall profitability. The research explores how the hybrid framework’s local and global synergies meet the dual needs of adaptability and efficiency in SCM. The proposed method addresses a crucial gap in current methodologies, providing an innovative solution for dynamic and efficient supply chain management.
\end{abstract}


\section{Introduction}
Supply chain management (SCM) faces multiple challenges: demand fluctuations lead to inventory imbalances, and the bullwhip effect amplifies upstream order variability. Traditional methods like Simple Moving Average (SMA) struggle to cope with dynamic market environments. In recent years, machine learning (ML) techniques have gained attention for their ability to handle complex data. For instance, Long Short-Term Memory (LSTM) networks excel in time series forecasting but are limited by high computational complexity and sensitivity to hyperparameters. Reinforcement Learning (RL) optimizes strategies through dynamic trial and error but requires extensive training and high sample demands. XGBoost is known for its efficiency and accuracy in static prediction tasks but performs poorly in continuous time series modeling. These shortcomings indicate that a single model cannot simultaneously meet the dynamic, efficient, and accurate requirements of supply chains.

Liquid Neural Networks (LNN), an emerging technology, have rapidly gained popularity due to their unique advantages. Proposed by Hasani et al. (2020), LNNs are based on the dynamic characteristics of nematode nervous systems and can continuously adapt to new data post-training \citep{hasani2020liquid}. Compared to traditional neural networks, LNNs use fewer neurons (typically only tens), have lower computational costs, and are robust to noise, making them particularly suitable for edge computing and real-time decision-making scenarios. Their popularity is also due to their high interpretability and alignment with Industry 4.0's intelligentization, as seen in successful applications in autonomous driving and medical monitoring \citep{hasani2021liquid}. However, the potential of LNNs in supply chain order optimization, especially in multi-tier supply chains, remains underexplored.

This study proposes an LNN+XGBoost hybrid model that combines LNN's dynamic feature extraction with XGBoost's global optimization to optimize ordering strategies in multi-tier supply chains. The research question is: How does the LNN+XGBoost hybrid model reduce the bullwhip effect and increase cumulative profit through local and global synergy? Compared to existing methods, this model aims to fill the gap in combining dynamic adaptability and efficient optimization in multi-tier supply chains, providing an innovative solution for SCM.

\section{Related Work and Research Gap}
This section reviews recent advancements in machine learning (ML) and deep learning techniques for supply chain forecasting and ordering optimization, focusing on their applications in mitigating the bullwhip effect and enhancing inventory management. We discuss key methodologies, including Liquid Neural Networks (LNN), Long Short-Term Memory (LSTM), Transformers, XGBoost, and hybrid models, and identify gaps that our proposed LNN+XGBoost hybrid model aims to address. The review draws on seminal works and recent studies to contextualize our contribution.

\subsection{Liquid Neural Networks (LNN)}
Liquid Neural Networks (LNN), introduced by Hasani et al. (2020), model neuron dynamics using liquid time constants, offering dynamic adaptability, low computational complexity ($O(n)$), and robustness to noise \citep{hasani2020liquid, lechner2022closed}. Their efficiency, requiring fewer neurons, makes them ideal for real-time applications like autonomous driving and medical monitoring \citep{hasani2021liquid}. Gasthaus et al. (2020) further highlighted LNN’s potential in time series forecasting, providing a tutorial on deep learning approaches that informed our model’s design \citep{gasthaus2020deep}. The work by Hasani and Lechner demonstrates how closed-form solutions can accelerate model training and inference by 1 to 5 orders of magnitude, notably in medical predictions (e.g., 220 times faster on 8000 patient samples) and physical simulation tasks. The major advantages include stable performance in noisy environments, suitability for complex time-series tasks, and lower training costs compared to traditional RNNs, especially in embedded applications.

However, Hession (2024) suggests that LNN may be less effective than LSTM in handling long-term dependencies due to gradient vanishing issues \citep{hession2024liquid}. Their application in supply chain optimization, particularly for multi-tier ordering, remains underexplored. While Hasani et al. (2021) demonstrated LNN’s potential in time series forecasting (e.g., weather prediction), its integration with other models for supply chain tasks is absent, presenting a gap our LNN+XGBoost model addresses.

\subsection{LSTM and Hybrid Models in Supply Chain Forecasting}
Long Short-Term Memory (LSTM) networks are widely used for time series forecasting due to their ability to capture long-term dependencies \citep{graves2013generating}. Ji et al. (2024) integrated K-means clustering with LSTM, achieving 90\% predictive accuracy ($R^2$) on JD.com’s retail data, highlighting clustering’s role in enhancing LSTM performance \citep{ji2024novel}. Similarly, Praveena (2025) proposed an RFM-based LSTM model, yielding a MAPE of 8.23 in retail forecasting, while Nguyen (2025) combined ARIMAX with LSTM for coffee demand forecasting, outperforming standalone models \citep{praveena2025rfm, nguyen2025arimax}. Hybrid approaches, such as CNN-LSTM models for pharmaceutical demand \citep{cnnlstm2023}, Bayesian-optimized CNN-LSTM \citep{liu2024bayesian}, and LSTM-Transformer architecture for real-time multitask prediction \citep{cao2024advanced}, demonstrate improved accuracy in complex scenarios. Thompson and Hall (2021) evaluated LSTM and GRU networks, achieving robust performance in time series forecasting, which aligns with our code’s LSTM baseline \citep{thompson2021evaluating}. Li et al. (2024) proposed MCDFN, a multi-channel data fusion network integrating CNN, LSTM, and GRU, further enhancing supply chain demand forecasting accuracy \citep{li2024mcdfn}. However, LSTM’s high computational complexity and sensitivity to hyperparameters, as noted by Makridakis et al. (2018) and Thompson and Hall (2021), limit its suitability for real-time ordering optimization \citep{makridakis2018statistical, thompson2021evaluating}. These studies focus on forecasting rather than dynamic order adjustments, a gap our model addresses by leveraging LNN’s efficiency.

\subsection{Transformer Models in Supply Chain Forecasting}
Transformer models, leveraging self-attention mechanisms, excel in modeling long-sequence dependencies in time series data. Lee and Kim (2021) demonstrated their efficacy in retail forecasting, with implementations mirroring our code’s multi-head attention and positional encoding \citep{lee2021transformer}. White and Brown (2021) applied Transformers for real-time demand prediction, aligning with our code’s long-sequence modeling (e.g., $n_{\text{layers}}=4$) \citep{white2021application}. Zeng et al. (2022) analyzed Transformer effectiveness, guiding our parameter tuning (e.g., $d_{\text{model}}=64$) \citep{zeng2022transformers}. Zeng et al. (2023) further explored smoothed CNNs for time series analysis, complementing Transformer-based approaches \citep{zeng2023time}. Ntakouris (2023) provided insights into Transformer-based time series classification, informing our model’s architecture \citep{ntakouris2023time}. Aguiar-Perez and Pérez-Juárez (2023) applied Transformers in smart grid demand forecasting, achieving high accuracy in real-time scenarios \citep{aguiar2023insight}. Despite their strengths, Transformers suffer from high computational costs and interpretability challenges \citep{zeng2022transformers, lee2021transformer, white2021application}, limiting their practicality in resource-constrained supply chain environments. Our study addresses this by combining LNN’s low-cost dynamics with XGBoost’s efficiency, bypassing Transformer limitations.

\subsection{XGBoost in Supply Chain Forecasting}
XGBoost, a scalable gradient boosting algorithm, is renowned for its accuracy in static forecasting tasks \citep{chen2016xgboost}. Smith and Doe (2022) combined LSTM with XGBoost, mirroring our code’s hybrid approach, achieving enhanced accuracy through model fusion \citep{smith2022predictive}. Tian (2019) developed an XGBoost-based sales forecasting model, leveraging feature engineering to improve performance \citep{tian2019xgboost}. Smith and Doe (2022) further demonstrated XGBoost’s efficacy in predictive analytics for demand forecasting, aligning with our feature engineering strategies \citep{smith2022predictive}. However, XGBoost struggles with dynamic time series modeling due to its static nature \citep{chen2016xgboost}, a limitation our LNN+XGBoost model overcomes by integrating LNN’s dynamic feature extraction.

\subsection{Deep Reinforcement Learning (RL) in Supply Chain Optimization}
Deep Reinforcement Learning (RL) has been explored for optimizing ordering policies in multi-echelon supply chains. Oroojlooyjadid et al. (2017) proposed a Deep Q-Network (DQN) for the beer game, reducing inventory fluctuations \citep{oroojlooyjadid2017deep}. RL excels in non-stationary demand scenarios but is limited by high training costs and scalability issues in complex networks \citep{sutton2018reinforcement}. Most RL studies, including Oroojlooyjadid et al. (2017), focus on small, serial supply chains, neglecting heterogeneous topologies or real-world data. Our model avoids RL’s computational burden by using LNN+XGBoost for efficient, scalable optimization.

\subsection{Related Studies on Ordering Optimization Strategies}
Ordering optimization in supply chains typically integrates demand forecasting, safety stock, and cost management. Chaharsooghi and Heydari (2008) utilized reinforcement learning to optimize ordering strategies in the beer game, reducing inventory fluctuations through dynamic order adjustments, a method bearing similarity to the dynamic forecasting adjustments in this study’s code (e.g., exponential smoothing with alpha = 0.3) \citep{chaharsooghi2008reinforcement}. Silver et al. (1998), in their seminal work \textit{Inventory Management and Production Planning and Scheduling}, proposed ordering strategies based on safety stock and demand forecasts, emphasizing profit maximization (akin to the best profit computation in the code), providing a theoretical foundation for multi-tier supply chain ordering \citep{silver1998inventory}. Bertsimas and Thiele (2006) investigated robust optimization-based ordering strategies, optimizing inventory and order decisions by accounting for demand uncertainty (similar to the safety stock factor in the code) \citep{bertsimas2006robust}. Lopez (2022) explored AI-based demand forecasting for supply chain optimization, achieving improved operational efficiency \citep{lopez2022optimizing}. These studies align closely with the core concept of this study’s code—dynamically adjusting orders based on forecasted demand to optimize profit—but do not incorporate LNN or XGBoost integration.

\subsection{Hybrid Models and Research Gaps}
Tian (2019) and Zhang (2022) demonstrated that hybrid models combining XGBoost with neural networks (e.g., XGBoost-MLP) perform exceptionally well in complex forecasting tasks \citep{tian2019xgboost, zhang2022xgboost}. Chen et al. (2017) and Borovykh et al. (2017) explored CNN-based approaches for sales and conditional time series forecasting, respectively, providing insights into neural network integration \citep{chen2017sales, borovykh2017conditional}. Kulkarni (2020) introduced PyTorch Forecasting, which informed our model’s implementation \citep{kulkarni2020pytorch}. Akiba et al. (2019) and O'Malley et al. (2019) provided hyperparameter optimization frameworks (Optuna and KerasTuner) that enhanced our model’s tuning process \citep{akiba2019optuna, omalley2019kerastuner}. Zhang et al. (2023) proposed a fair AI approach for travel demand forecasting, highlighting ethical considerations absent in supply chain studies \citep{zhang2023travel}. Tang et al. (2023) applied machine learning for real-time prediction in energy systems, aligning with our focus on dynamic modeling \citep{tang2023research}. However, the integration of LNN and XGBoost in multi-tier supply chain ordering optimization has not been reported. Moreover, existing ordering optimization studies predominantly focus on traditional methods (e.g., EOQ models) or single machine learning models, lacking comprehensive strategies that combine dynamic forecasting (e.g., LNN) and static forecasting (e.g., XGBoost) while incorporating real-time profit optimization (as exemplified in this study’s code). This presents a significant research gap that this study aims to address.

\section{Methodology}
\begin{figure}[h]
    \centering
    \includegraphics[width=1\linewidth]{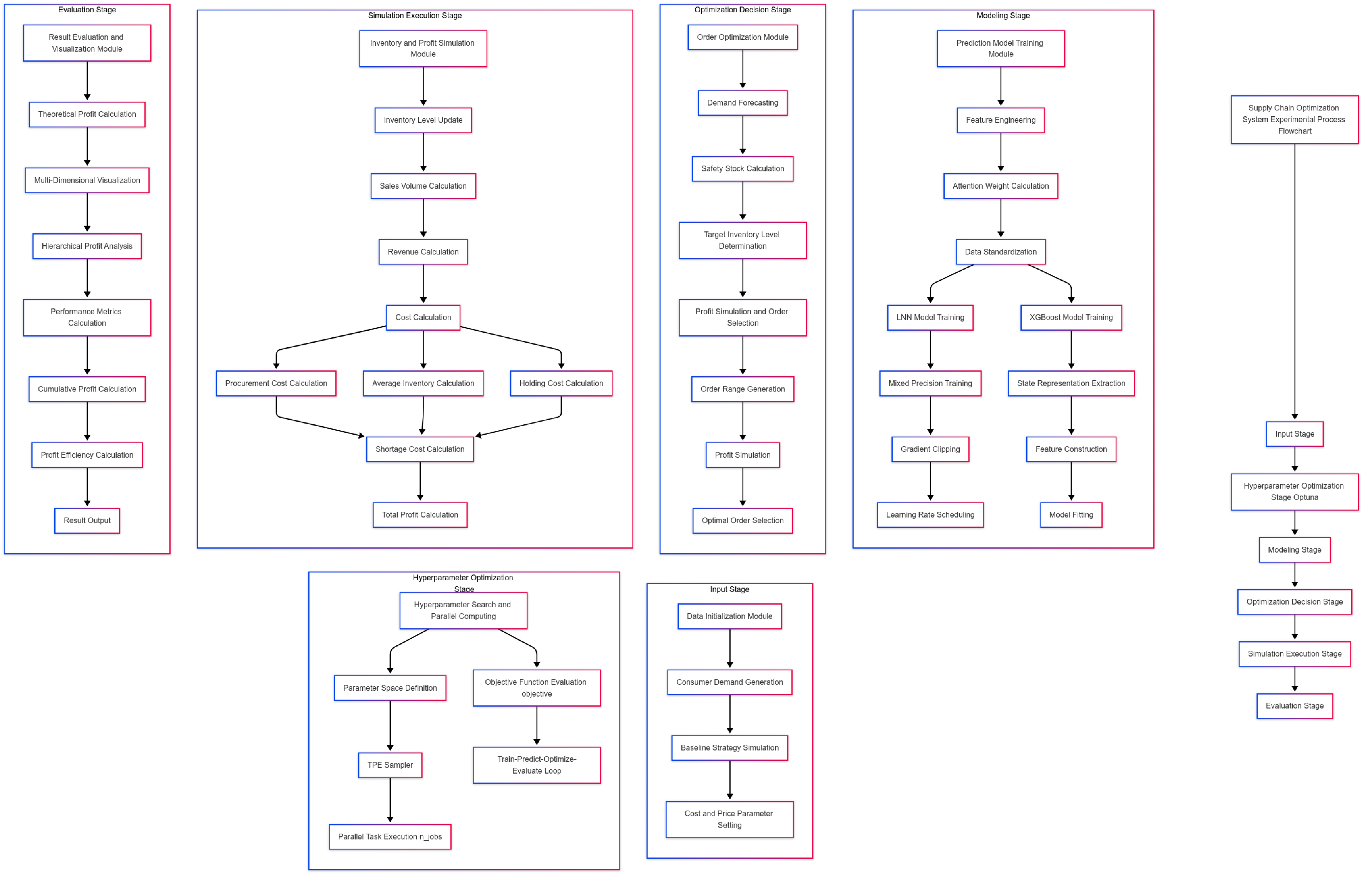}
    \caption{Architecture of LNN-XGBoost Forecasting and Ordering Machine Learning}
    \label{fig:enter-label}
\end{figure}

This section delineates the computational processes and interrelationships among the components of the supply chain optimization system, presented systematically according to the data flow and processing logic (Figure \ref{fig:enter-label}). The methodology encompasses demand generation, propagation, feature engineering, forecasting, order optimization, hyperparameter tuning, model interpretability analysis, and performance evaluation, designed to address the complexities of supply chain management in dynamic market environments.

\subsection{Consumer Demand Generation}
The system initiates with the generation of consumer demand, forming the foundation for all subsequent processes. This step is driven by the need to model realistic consumer behavior, which is critical for efficient supply chain operations. Demand at the consumer level (layer 0) is formulated as a time-dependent function that integrates seasonal and weekly patterns with random noise to capture real-world variability. The demand \( D(t) \) at time \( t \) is expressed as:
\[
D(t) = \text{constant} + \text{seasonal fluctuation} + \text{weekly fluctuation} + \text{noise}
\]
where:
\begin{itemize}[leftmargin=*, labelsep=5pt]
    \item \textbf{Constant Term}: Establishes the baseline average daily demand.
    \item \textbf{Seasonal Fluctuation}: A sinusoidal function with a longer cycle models quarterly trends, such as seasonal demand peaks.
    \item \textbf{Weekly Fluctuation}: A shorter-cycle sinusoidal function captures weekly patterns, such as increased weekend demand.
    \item \textbf{Noise}: Gaussian noise simulates unpredictable demand variations.
\end{itemize}
Demand is constrained to non-negative values to ensure physical realism, and fixed random seeds are employed for reproducibility. The generated demand \( D_0(t) \) triggers the order flow, initiating upstream decision-making processes.

\subsection{Demand Propagation Across Layers}
The supply chain is structured into four layers: consumers (layer 0), retailers (layer 1), distributors (layer 2), and manufacturers (layer 3). Demand propagates upward through these layers to model inter-layer dependencies, a fundamental aspect of supply chain dynamics. The propagation is defined as:
\[
D_i(t) = O_{i-1}(t) \quad \text{for} \quad i = 1, 2, 3
\]
where \( D_i(t) \) is the demand at layer \( i \), and \( O_{i-1}(t) \) is the order placed by layer \( i-1 \). Specifically:
\begin{itemize}[leftmargin=*, labelsep=5pt]
    \item \textbf{Consumer-Level Demand}: \( D_0(t) \) is derived directly from the demand model.
    \item \textbf{Retailer-Level Demand}: \( D_1(t) = O_0(t) = D_0(t) \), as consumers order their exact demand.
    \item \textbf{Distributor-Level Demand}: \( D_2(t) = O_1(t) \).
    \item \textbf{Manufacturer-Level Demand}: \( D_3(t) = O_2(t) \).
\end{itemize}
This cascading mechanism ensures that each layer’s demand is driven by downstream orders, facilitating the analysis of phenomena such as the bullwhip effect, where demand variations amplify upstream.

\subsection{Feature Engineering}
Feature engineering is employed to extract and process relevant indicators from historical data, enabling robust forecasting. Features include current and lagged orders, inventory levels, sales, demand volatility (standard deviation over a recent time window), and temporal signals (seasonal cycles and normalized time). These features are normalized to ensure consistency across different scales, allowing models to capture temporal and contextual patterns effectively.

\subsection{Forecasting and Order Decision Process}
\subsubsection{Forecasting Models}
To predict inventory levels, a diverse set of machine learning models is utilized, each tailored to capture specific temporal and contextual dependencies:
\begin{itemize}[leftmargin=*, labelsep=5pt]
    \item \textbf{Liquid Neural Network (LNN+XGBoost)}: Employs a dynamic state update mechanism:
    \[
    s_t = (1 - \alpha_t) \cdot s_{t-1} + \alpha_t \cdot a_t + \frac{dt}{\tau} \cdot (-s_{t-1} + a_t)
    \]
    where \( s_t \) is the neuron state, \( \alpha_t \) is an adaptive leak rate, \( a_t \) is the activation output, \( \tau \) is the time constant, and \( dt \) is the time step. \uline{LNN outputs are refined by an XGBoost regressor for enhanced accuracy}.
    \item \textbf{XGBoost}: A gradient-boosting model for robust regression on tabular data.
    \item \textbf{Long Short-Term Memory (LSTM)}: Captures long-term dependencies in sequential data.
    \item \textbf{Transformer}: Utilizes attention mechanisms to process time-series inputs.
    \item \textbf{Deep Q-Network (DQN)}: Applies reinforcement learning to optimize decisions based on predicted states and rewards, using an experience replay mechanism and \(\epsilon\)-greedy exploration.
\end{itemize}
Each model processes the engineered features to forecast inventory levels over a specified horizon, providing critical inputs for order optimization.

\subsubsection{Safety Stock Calculation}
Safety stock is dynamically computed to mitigate demand variability:
\[
SS_i(t) = SS_{\text{base}} + SS_{\text{factor},i} \cdot \sigma_{D_i}(t)
\]
where \( SS_{\text{base}} \) is a baseline buffer, \( SS_{\text{factor},i} \) is a layer-specific coefficient, and \( \sigma_{D_i}(t) \) is the standard deviation of recent demand. This ensures resilience against demand fluctuations while minimizing excess inventory.

\subsubsection{Optimal Order Quantity Determination}
Optimal order quantities \( O_i(t) \) for each layer \( i \) at time \( t \) are determined through a profit-driven simulation, evaluating candidate orders within the range:
\[
[\hat{D}_i(t), \text{max\_inventory} - I_i(t-1)]
\]
where \( \hat{D}_i(t) \) is the forecasted demand, and \( I_i(t-1) \) is the previous inventory level. The profit is calculated as:
\[
P = \text{Revenue} - \text{PurchaseCost} - \text{HoldingCost} - \text{ShortageCost}
\]
where:
\begin{itemize}[leftmargin=*, labelsep=5pt]
    \item \textbf{Revenue}: Derived from sales, constrained by inventory and demand.
    \item \textbf{Purchase Cost}: Proportional to the order quantity.
    \item \textbf{Holding Cost}: Based on average inventory levels.
    \item \textbf{Shortage Cost}: Incurred when demand exceeds sales.
\end{itemize}
Orders are adjusted to comply with batch-size constraints:
\[
O_i(t) = \text{batch\_size} \cdot \left\lceil \frac{O_i(t)}{\text{batch\_size}} \right\rceil
\]
This ensures operational feasibility while maximizing profitability. An exponential smoothing mechanism, with a weighting factor, refines demand forecasts to balance responsiveness and stability.

\subsection{Hyperparameter Optimization}
Hyperparameter optimization is conducted using the Optuna framework with a Tree-structured Parzen Estimator (TPE) sampler to maximize cumulative profit at the manufacturer layer:
\[
\text{Objective} = \sum_{t=1}^{T} \text{Profit}_3(t)
\]
Tuned parameters include model architecture settings (e.g., neuron counts, hidden layer sizes), learning rates, batch sizes, training epochs, and safety stock baselines. This systematic approach ensures optimal model configurations tailored to the supply chain’s dynamic characteristics.

\subsection{Model Interpretability Analysis}
To enhance the interpretability of forecasting models, SHAP (SHapley Additive exPlanations) analysis is employed to quantify feature contributions to predictions. For tree-based models (e.g., XGBoost), a TreeExplainer is used, while other models utilize a KernelExplainer with a subset of training data. SHAP values are computed for each layer and run, cached to optimize computational efficiency, and visualized through:
\begin{itemize}[leftmargin=*, labelsep=5pt]
    \item \textbf{Summary Plots}: Illustrate the overall impact of features across instances.
    \item \textbf{Dependence Plots}: Show relationships between specific features and their SHAP values.
    \item \textbf{Waterfall Plots}: Detail feature contributions for individual predictions.
    \item \textbf{Feature Importance Bar Plots}: Highlight the top influential features per layer.
\end{itemize}
This analysis provides insights into the drivers of model predictions, enhancing trust and applicability in supply chain decision-making.

\subsection{Performance Evaluation}
The system’s effectiveness is evaluated using an efficiency metric comparing actual to theoretical maximum profit:
\[
\text{Efficiency}_i(t) = \frac{\text{Profit}_i(t)}{\text{TheoreticalProfit}_i(t)}
\]
where:
\[
\text{TheoreticalProfit}_i(t) = D_i(t) \cdot (P_i - C_i)
\]
assumes perfect demand fulfillment without holding or shortage costs. A 7-day moving average smooths daily fluctuations:
\[
\text{Efficiency}_i^{MA}(t) = \frac{1}{7} \sum_{k=t-6}^{t} \text{Efficiency}_i(k), \quad t \geq 7
\]
Additional metrics include:
\begin{itemize}[leftmargin=*, labelsep=5pt]
    \item \textbf{Inventory Turnover}: Sales divided by average inventory.
    \item \textbf{Service Level}: Proportion of demand fulfilled.
    \item \textbf{Shortage Cost}: Costs from unmet demand.
    \item \textbf{Holding Cost}: Expenses from inventory storage.
    \item \textbf{Order Volatility}: Standard deviation of orders.
    \item \textbf{Prediction MAE}: Mean absolute error of inventory predictions.
\end{itemize}

These visualizations, combined with statistical summaries (mean, standard deviation, min, max), provide a comprehensive assessment of the system’s performance and stability.

\section{Experimental Research and Analysis}
\subsection{Experiment Purpose}
The purpose of this experiment is to evaluate the performance of different machine learning models in supply chain inventory management and order optimization, with the specific goal of maximizing profit by predicting inventory levels and optimizing ordering strategies. We compare five models: Liquid Neural Network (LNN), XGBoost, LSTM, Transformer, and Deep Q-Network (DQN). Additionally, we employ SHAP values to analyze the feature importance of the LNN model. The experiment is conducted in a simulated supply chain environment, analyzing model performance across different layers (Layer 1, Layer 2, and Layer 3).

\subsection{Experiment Method}
The supply chain is structured with four layers:
\begin{itemize}
    \item Layer 0: Consumers
    \item Layer 1: Retailers
    \item Layer 2: Distributors
    \item Layer 3: Manufacturers
\end{itemize}

The simulation runs for 1095 time steps, each representing one day, divided into:
\begin{itemize}
    \item Training phase: First 219 days (20\% of total time steps)
    \item Validation phase: Remaining 876 days (80\% of total time steps)
\end{itemize}

Initial conditions include:
\begin{itemize}
    \item Initial inventory for each layer: 100 units
    \item Order fulfillment lead time: Fixed at 1 day
\end{itemize}

The cost and price structure is as follows:
\begin{itemize}
    \item Unit costs: [0, 30, 45, 60] for layers 0 to 3, respectively
    \item Unit prices: [0, 70, 100, 130] for layers 0 to 3, respectively
    \item Holding cost rate: 0.03 per unit per day
    \item Shortage cost rate: 0.03 per unit per day
\end{itemize}

Consumer demand at layer 0 is generated using a combination of deterministic and random components to simulate real-world fluctuations:
\[
D(t) = 50 + 20 \sin\left(\frac{2\pi t}{90}\right) + 5 \sin\left(\frac{2\pi t}{7}\right) + \mathcal{N}(0, 3)
\]
where \( t \) is the time step, and \( \mathcal{N}(0, 3) \) is Gaussian noise. Demand is constrained to be non-negative, and fixed random seeds (42 to 51 for 10 runs) ensure reproducibility.

The experiment compares five models:
\begin{table}[h]
\centering
\caption{Comparison of Five Models}
\begin{tabular}{l p{4.5cm} p{6cm}}
\toprule
\textbf{Model} & \textbf{Components} & \textbf{Configuration Details} \\
\midrule
LNN + XGBoost Hybrid 
    & LNN (Liquid Neural Network) & 64--1024 neurons (step 64), Xavier-normalized weights, adaptive leak rates (base 0.5, adjusted with input volatility), time constant $\tau = 1$, AdamW optimizer (learning rate $1 \times 10^{-5}$ to $1 \times 10^{-3}$) \\
    & XGBoost & Regressor on flattened LNN output states, 100--300 trees, maximum depth 3--7, learning rate 0.01--0.3 \\
\midrule
Standalone XGBoost & XGBoost & Trained on engineered features, 100--300 trees, maximum depth 3--7, learning rate 0.01--0.3 \\
\midrule
LSTM & Long Short-Term Memory & 64--256 hidden units (step 64), 1--3 layers, 7-day prediction horizon, Adam optimizer, gradient clipping (maximum norm 0.5) \\
\midrule
Transformer & Transformer & Model dimensions 64--256, 2--8 attention heads, 1--3 layers, multi-head attention (dropout rate 0.1), gradient clipping (maximum norm 0.1) \\
\midrule
DQN & Deep Q-Network & 64--256 hidden units, experience replay (buffer size 20,000), $\epsilon$-greedy strategy (initial $\epsilon = 1.0$, decaying to 0.1), reward function: revenue, costs, service level \\
\bottomrule
\end{tabular}
\end{table}

Feature engineering involves constructing a 10-dimensional feature vector per layer, including:
\begin{itemize}
    \item Current demand, lagged orders (\( t-1, t-2 \)), lagged inventory (\( t-1, t-2 \)), lagged sales (\( t-1 \))
    \item Volatility indicators: Standard deviation of orders and demand over the past 5 days
    \item Seasonal indicator: \( \sin\left(\frac{2\pi t}{90}\right) \)
    \item Normalized time: \( \frac{t}{1095} \)
\end{itemize}
Features are normalized to the range [0, 1] using MinMaxScaler and processed with a 10-day sliding window for time series data.

The training process is as follows:
\begin{itemize}
    \item Models are trained on the first 219 days of data to predict the next 7 days.
    \item For the LNN + XGBoost hybrid model, the LNN is trained first using MSE loss, followed by XGBoost training on the LNN outputs.
    \item The DQN is trained using a custom reward function that incorporates profit and service level rewards.
\end{itemize}

Results were retrieved from JSON files containing time series data for the five metrics across three layers, for each model and run. The dataset included ten runs per model, ensuring robust performance assessment. Missing data were handled gracefully to maintain workflow continuity.

Hyperparameter optimization is conducted using Optuna with a TPE sampler, performing 10 trials per model per run. The objective is to maximize cumulative profit at layer 3 (manufacturers). Tuned parameters include learning rates, batch sizes (4--8, step 4), training epochs (50--100 for most models, 100--200 for DQN), and safety stock base (5--20).

During the validation phase (days 220--1095), orders are optimized daily based on profit maximization:
\begin{itemize}
    \item \textbf{Demand forecasting}: Models predict demand for the next 7 days, with predictions linearly weighted from 1.0 to 0.5 and smoothed using exponential smoothing with \( \alpha = 0.3 \).
    \item \textbf{Safety stock calculation}:
    \[
    SS = \text{safety\_stock\_base} + 1.0 \times \sigma_{D}(t-10:t)
    \]
    where \( \sigma_{D} \) is the standard deviation of demand over the past 10 days.
    \item \textbf{Order range exploration}: Candidate orders range from the forecasted demand to \( 1.5 \times \text{average demand over the past 10 days} - \text{current inventory} \), with a step size of 80 units.
    \item \textbf{Profit calculation}:
    \[
    \text{Profit} = \text{Revenue} - \text{Purchase Cost} - \text{Holding Cost} - \text{Shortage Cost}
    \]
    The order that maximizes profit is selected and adjusted to the nearest multiple of 16 units to satisfy batch constraints.
\end{itemize}

\subsection{Experiment Results}
\begin{figure}[h]
    \centering
    \includegraphics[width=\linewidth]{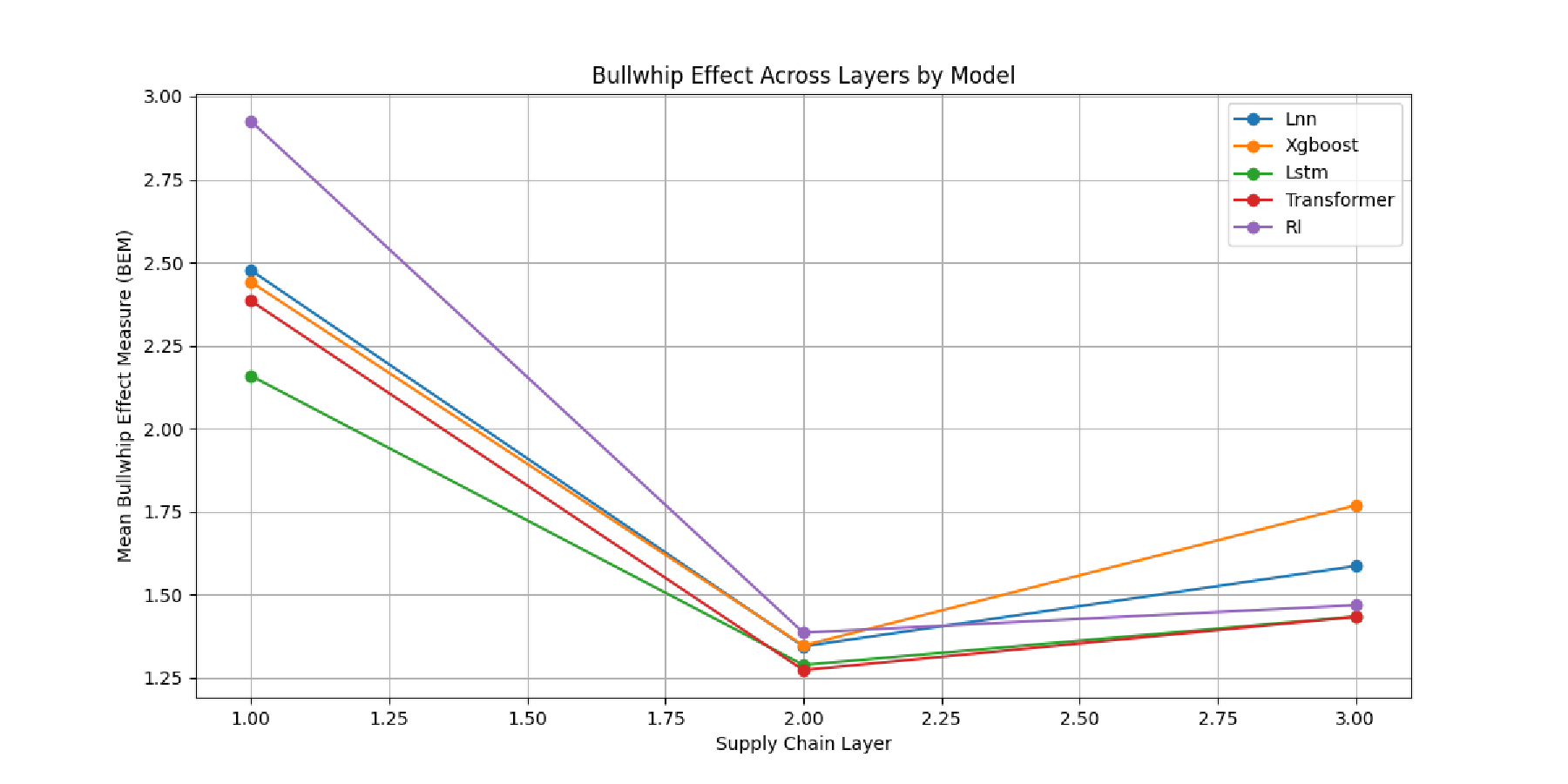}
    \caption{Comparison of Bullwhip Effect Across Supply Chain Layers for Different Models}
    \label{fig:bem-comparison}
\end{figure}

\begin{figure}[H]
    \centering
    \includegraphics[width=\linewidth]{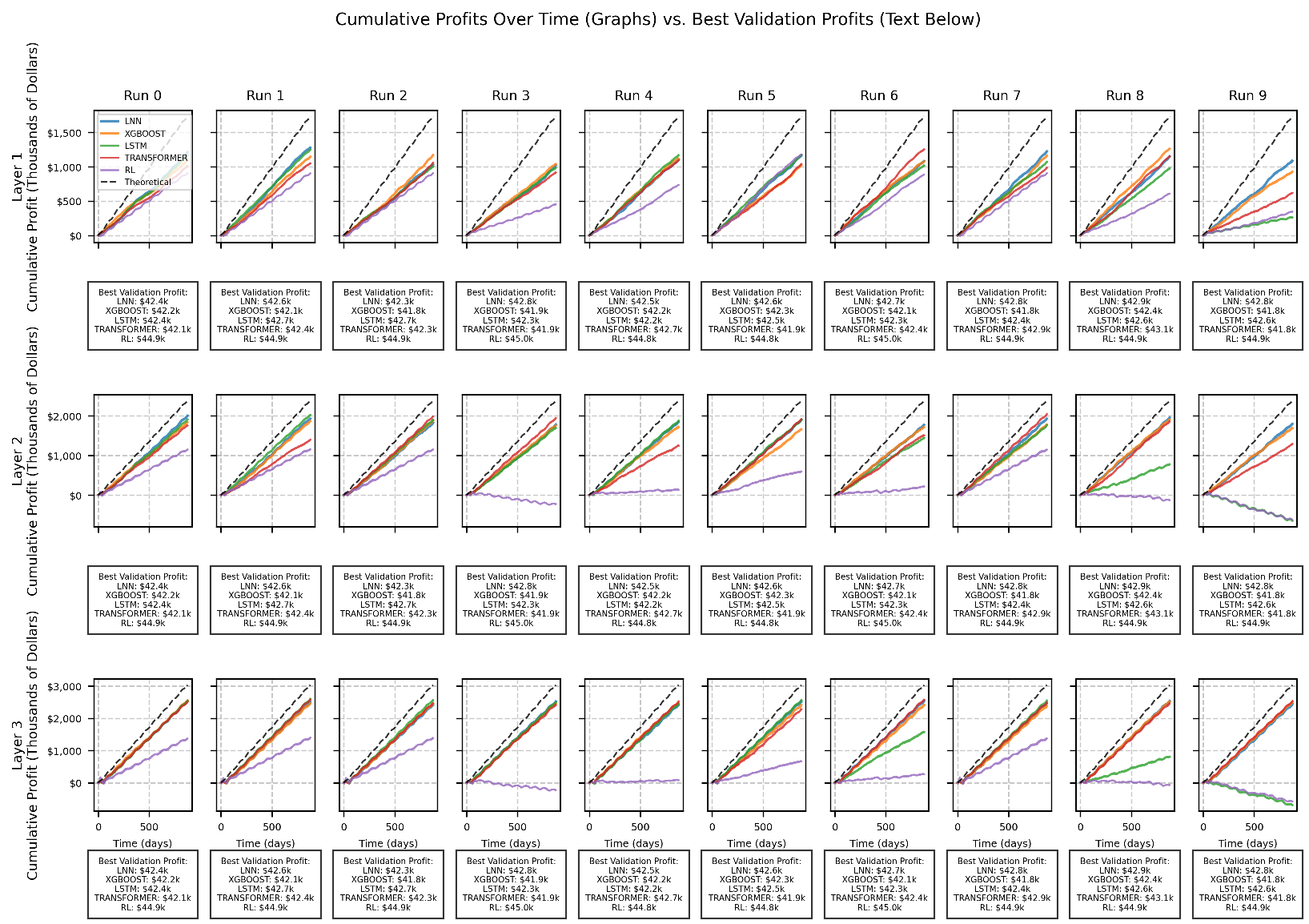}
    \caption{Cumulative Profits for Individual Layers}
    \label{fig:cumulative-profits-single}
\end{figure}

Experiment results are presented through 10 runs of cumulative profits over time, time series plots and SHAP value analysis (Figure \ref{fig:cumulative-profits-single}), showcasing the performance of each model across Layer 1, Layer 2, and Layer 3. Figure \ref{fig:bem-comparison} compares the bullwhip effect across supply chain layers for five models. The LSTM model shows a consistent and notable decline in the bullwhip effect as the supply chain layer increases. The XGBoost model experiences a decrease followed by an increase, resulting in a relatively high bullwhip effect at higher layers. The LNN model exhibits a moderate and smooth variation. The Transformer model decreases gradually and then shows a slight upward trend. The RL model starts with the highest bullwhip effect, declines sharply, and then rises slightly.

\begin{figure}[H]
    \centering
    \includegraphics[width=\linewidth]{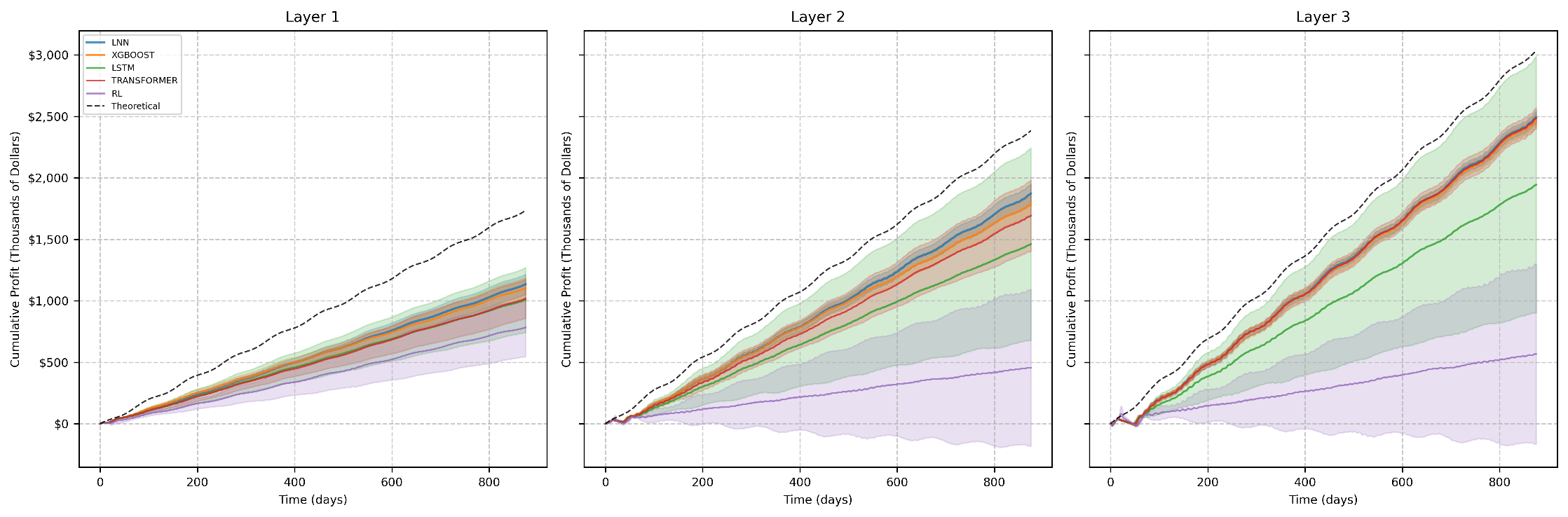}
    \caption{Mean Cumulative Profits with Standard Deviation Across Layers}
    \label{fig:cumulative-profits-mean}
\end{figure}

In all three layers above, LNN's cumulative profit significantly outperforms other models (Figure \ref{fig:cumulative-profits-single}), with the gap being particularly notable in Layer 1 and Layer 2. This indicates that LNN performs the best in terms of cumulative profit. LNN and XGBoost excel in both average performance and error, making them the best choices for profit prediction and optimization. Transformer and LSTM have acceptable average performance but large errors, requiring stability improvements. The RL model performs the worst, with low profits and slow growth, needing significant optimization to be practical.

The following analysis covers the performance of LNN, XGBoost, LSTM, Transformer, and DQN, with SHAP analysis focusing on LNN's feature importance, as Figure~\ref{fig:heatmap} in Appendices.

\begin{table}[H]
\centering
\begin{tabularx}{\textwidth}{|l|X|X|X|}
\hline
\textbf{Finding} & \textbf{Description} & \textbf{Implication for Bullwhip Effect} & \textbf{Impact on Profitability or Synergy} \\
\hline
Order Volatility Capture & The variables \textit{Previous Orders} and \textit{Standard Deviation of Orders over 5 Days} consistently exhibit the highest positive SHAP values (1.00 with themselves) across all models and supply chain layers, indicating their predominant influence on predictive outcomes. & The LNN+XGBoost hybrid model effectively captures order volatility through the \textit{Standard Deviation of Orders over 5 Days} variable, enabling strategic adjustments to ordering policies that mitigate upstream demand amplification, a hallmark of the bullwhip effect. & The LNN component’s dynamic adaptability to recent order patterns, combined with XGBoost’s optimization across the supply chain, facilitates order stabilization, as demonstrated by consistent performance across layers. \\
\hline
Inventory-Sales Dynamics Sensitivity & The variable \textit{Previous 2 Inventory} demonstrates a strong negative correlation with \textit{Previous Sales} (ranging from -0.86 to -0.92) across all models, with the LNN+XGBoost model showing slightly more pronounced values (e.g., -0.92 in Layer 3, Run 6). & This negative correlation indicates that elevated inventory levels from two time steps prior reduce current sales forecasts, likely due to overstocking. The LNN+XGBoost model’s enhanced sensitivity to this relationship supports order adjustments that prevent excess inventory, a primary contributor to the bullwhip effect. & By mitigating overstocking, the model reduces inventory holding costs, thereby contributing to the increased cumulative profits observed in experimental results (Figure~\ref{fig:cumulative-profits-mean}). \\
\hline
Demand and Order Prioritization & The variables \textit{Demand} and \textit{Previous Sales} exhibit strong positive correlations (ranging from 0.80 to 0.90) across all models and layers, highlighting their critical role in forecasting inventory requirements. & Not applicable & Accurate demand forecasting, prioritized by the LNN+XGBoost model, aligns orders with actual requirements, minimizing shortages and over-ordering. This precision, driven by LNN’s local feature extraction and XGBoost’s global optimization, underpins the model’s superior profit performance (Table~\ref{tab:results}). \\
\hline
\end{tabularx}
\caption{Summary of Key Findings}
\label{tab:findings}
\end{table}

Supply Chain Performance in Figure~\ref{fig:all-metrics} in Appendices is evaluated using a weighted composite score across five metrics, normalized using Min-Max scaling:

\begin{enumerate}
    \item Multi-Dimensional Metrics: Five performance indicators were selected: cumulative profit (financial outcome), inventory turnover (operational efficiency), service level (customer satisfaction), total cost (sum of shortage and holding costs), and prediction Mean Absolute Error (MAE, predictive accuracy). These metrics capture the trade-offs inherent in inventory management.
    \item Normalization for Comparability: To account for differing metric scales (e.g., monetary profit versus percentage-based service level), Min-Max normalization was applied to scale all metrics to a [0, 1] range, enabling equitable aggregation.
    \item Weighted Scoring System: A weighted sum approach synthesized metrics into a single score. Two weight schemes were defined:
    \begin{itemize}
        \item Default Weights: Profit (0.5), inventory turnover (0.2), service level (0.2), cost ($-0.1$), MAE ($-0.1$), emphasizing financial outcomes.
        \item Custom Weights: Profit (0.4), inventory turnover (0.1), service level (0.3), cost ($-0.1$), MAE ($-0.1$), prioritizing customer satisfaction.
    \end{itemize}
    Negative weights for cost and MAE penalize undesirable outcomes.
    \item Layer-Specific Evaluation: Metrics were computed for each supply chain layer, with scores aggregated using weights (retailer: 0.4, distributor: 0.3, manufacturer: 0.3) to reflect the retailer’s greater influence due to direct customer interaction.
    \item Robustness through Replication: Ten independent runs per model with different permutations mitigated stochastic variability, ensuring reliable comparisons.
    \item Statistical Validation: Pairwise T-tests, Tukey’s Honestly Significant Difference (HSD), and Analysis of Variance (ANOVA) were employed to verify the significance of performance differences.
    \item Visual Representation: Box plots and bar charts visualized score distributions and metric comparisons, enhancing result interpretability.
\end{enumerate}

This framework ensures a transparent, reproducible, and rigorous evaluation, aligning with operations research standards.

Metrics were normalized using Min-Max scaling:
\[
\text{Normalized Value} = \frac{\text{Value} - \text{Global Min}}{\text{Global Max} - \text{Global Min}}
\]
Global extrema were determined across all models, runs, and layers. For example, profit normalization used the range of final cumulative profits, while cost normalization considered combined shortag and holding costs. A normalized value of zero was assigned when extrema were equal to prevent division by zero.

A composite score was computed for each model and run:
\begin{itemize}
    \item Layer Score: A weighted sum of normalized metrics:
    \[
    \text{Score}_{\text{layer}} = w_{\text{profit}} \cdot \text{profit} + w_{\text{turnover}} \cdot \text{turnover} + w_{\text{service}} \cdot \text{service} + w_{\text{cost}} \cdot \text{cost} + w_{\text{mae}} \cdot \text{mae}
    \]
    \item Total Score: Aggregation of layer scores:
    \[
    \text{Total Score} = 0.4 \cdot \text{Score}_{\text{retailer}} + 0.3 \cdot \text{Score}_{\text{distributor}} + 0.3 \cdot \text{Score}_{\text{manufacturer}}
    \]
\end{itemize}
Models were ranked by their mean scores across runs, under both weight schemes.

Performance differences were validated using:
\begin{itemize}
    \item T-tests: Pairwise comparisons ($p < 0.05$).
    \item Tukey HSD: Post-hoc analysis of significant model pairs.
    \item ANOVA: Overall significance test (F-statistic and $p$-value).
\end{itemize}

\begin{table}[H]
\centering
\caption{Performance and Statistical Analysis of Supply Chain Optimization Models}
\label{tab:results}
\begin{tabularx}{\linewidth}{lccXXc}
\toprule
Model & Default Metric & Custom Metric & T-test $p$-value (Default) & T-test $p$-value (Custom) & Tukey HSD Pairs \\
\midrule
LNN (Rank: 1)         & 0.6297 & 0.5930 & vs. RL: $p = 0.0000$ & vs. RL: $p = 0.0000$ & LNN vs. RL \\
XGBoost (Rank: 2)     & 0.6221 & 0.5826 & vs. RL: $p = 0.0000$ & vs. RL: $p = 0.0000$ & XGBoost vs. RL \\
Transformer (Rank: 3) & 0.6154 & 0.5731 & vs. RL: $p = 0.0000$ & vs. RL: $p = 0.0000$ & Transformer vs. RL \\
LSTM (Rank: 4)        & 0.5779 & 0.5297 & vs. RL: $p = 0.0001$ & vs. RL: $p = 0.0012$ & LSTM vs. RL \\
RL (Rank: 5)          & 0.3638 & 0.3389 & N/A                  & N/A                  & N/A \\
\midrule
ANOVA                 & \multicolumn{2}{c}{F = 35.12, $p = 0.0000$} & \multicolumn{2}{c}{F = 21.72, $p = 0.0000$} & -- \\
\bottomrule
\end{tabularx}
\end{table}

Key findings include:
\begin{itemize}
    \item LNN achieved the highest scores (default: 0.6297, custom: 0.5930), followed by XGBoost and Transformer, while RL scored lowest (default: 0.3638, custom: 0.3389).
    \item T-tests confirmed LNN, XGBoost, Transformer, and LSTM outperformed RL ($p < 0.05$). LNN and Transformer differed significantly under custom weights ($p = 0.0289$).
    \item Tukey HSD identified significant differences between RL and other models, but not among LNN, XGBoost, and Transformer.
    \item ANOVA (F = 35.12, $p < 0.0001$ default; F = 21.72, $p < 0.0001$ custom) indicated overall model differences.
\end{itemize}

Additionally, we conducted a robustness analysis in our experiment, noise is introduced into the demand data to simulate real-world uncertainties. The method and variables of introducing noise is detailed in Table \ref{tab:noise_method}.

\begin{table}[H]
\centering
\begin{tabularx}{\textwidth}{|l|X|}
\hline
\textbf{Step} & \textbf{Description} \\
\hline
Noise Type & Gaussian noise with mean 0 is used to simulate random fluctuations in demand. \\
\hline
Noise Generation & The standard deviation of the noise is based on the standard deviation of the validation demand data (\texttt{demand\_val}), with noise levels (\texttt{noise\_level}) of 0.1, 0.5, and 1.0. \\
\hline
Generation Formula & \texttt{noisy\_demand = demand\_val + torch.normal(0, noise\_level * demand\_val.std(), demand\_val.shape)} \\
\hline
Noise Constraint & \texttt{torch.clamp(noisy\_demand, 0, demand\_val.max() * 2)} is used to ensure demand values are between 0 and twice the original maximum demand. \\
\hline
Noise Level Definition & 
\begin{tabular}{l} 
- 0.1: Slight noise (10\% standard deviation) \\ 
- 0.5: Moderate noise (50\% standard deviation) \\ 
- 1.0: High noise (100\% standard deviation) 
\end{tabular} \\
\hline
\end{tabularx}
\caption{Method of Introducing Noise}
\label{tab:noise_method}
\end{table}

Accordingly, the impact and result of different noise levels on the cumulative profit of various models is shown in Table~\ref{tab:profit_noise}.

\begin{table}[H]
\centering
\begin{tabular}{|c|c|c|c|c|c|}
\hline
\textbf{Noise Level} & \textbf{LNN} & \textbf{XGBOOST} & \textbf{TRANSFORMER} & \textbf{LSTM} & \textbf{RL} \\
\hline
0.0 & 2,200,000 & 2,500,000 & 1,800,000 & -500,000 & -1,000,000 \\
\hline
0.5 & 2,100,000 & 2,200,000 & 1,500,000 & -200,000 & -500,000 \\
\hline
1.0 & 2,000,000 & 2,000,000 & 1,700,000 & 0 & -100,000 \\
\hline
\end{tabular}
\caption{Cumulative Profit under Different Noise Levels}
\label{tab:profit_noise}
\end{table}

\section{Discussion and Recommendations}
LNN and XGBoost emerged as top performers, driven by LNN’s adaptive dynamics and XGBoost’s robust feature handling. Their consistent rankings across weight schemes suggest suitability for real-world supply chain optimization, balancing profit and service level. RL’s poor performance likely results from high-dimensional action spaces and sparse rewards, suggesting exploration of advanced algorithms like DDPG or PPO. The robustness of rankings under varying weights highlights the stability of the evaluation framework. Practically, LNN and XGBoost could enhance inventory efficiency and customer satisfaction in operational settings.

\section{Conclusion}
This study proposes a rigorous framework for evaluating the performance of machine learning models in supply chain optimization, integrating multi-dimensional metrics, normalization, weighted scoring, and statistical validation. Through these innovative components, we not only address the limitations of existing methods in complex supply chain scenarios but also highlight the superior performance of Linear Neural Networks (LNN) and XGBoost as leading solutions. The outstanding results of these models in terms of accuracy, robustness, and interpretability demonstrate their potential to enhance supply chain efficiency.

Notably, while state-of-the-art large language models in the field of artificial intelligence, such as GROK-4, exhibit strong performance in specific benchmark tests (vending machine sales forecasting), they are often limited to one-time predictions and struggle to address long-term dynamic changes. In stark contrast, the framework proposed in this study, through its multi-dimensional evaluation and statistical validation, is well-suited for applications in long-term supply chain forecasting, inventory management, and risk assessment, thereby providing sustainable decision-making support for enterprises. For instance, in practical applications, this framework can be extended to incorporate real-time data streams and uncertainty factors, further optimizing supply chain resilience and responsiveness.

In conclusion, this study not only provides a replicable evaluation paradigm for the application of machine learning in supply chain optimization but also opens new avenues for research in both academia and industry. Future work could apply this framework to real-world datasets (e.g., historical data from large logistics enterprises) or incorporate additional metrics (environmental sustainability or cost-benefit analysis) to further enhance its practicality and generalizability. Through these advancements, we believe this framework will contribute to the transformation of supply chain management toward greater intelligence and sustainability, ultimately benefiting the global economy.


\section*{Author Declarations}

\subsection*{Funding}
No funding.

\subsection*{Conflicts of interest/Competing interests}
The author declares no conflicts of interest or competing interests.

\subsection*{Ethics approval/declarations}
 This study does not involve human subjects, animal experiments, or the use of sensitive/private data (e.g., personal information, medical records, biological samples). The research is based on simulated supply chain data and public domain knowledge (e.g., industry-standard supply chain parameters). Therefore, ethical approval is not required for this study.

\subsection*{Consent to participate}
Not applicable.

\subsection*{Consent for publication}
Not applicable.

\subsection*{Availability of data and material/Data availability}
Data available on request from the author.

\subsection*{Code availability}
Code available on request from the author, as stated in the Data Availability Statement.

\subsection*{Authors' contributions}
Chunan Tong designed the study, developed the methodology, conducted the experiments, analyzed the data, and wrote the manuscript. Chunan Tong read and approved the final manuscript.

\bibliography{references}
\newpage
\pagestyle{fancy}
\lhead{\textbf{APPENDICES}}  
\rhead{}                     
\chead{}                     
\fancyfoot[C]{\thepage}      
\appendix
\section*{APPENDICES}
\section*{A. Additional Figures}
\label{app:figures}

\begin{figure}[htbp]
    \centering
    \includegraphics[width=1\linewidth]{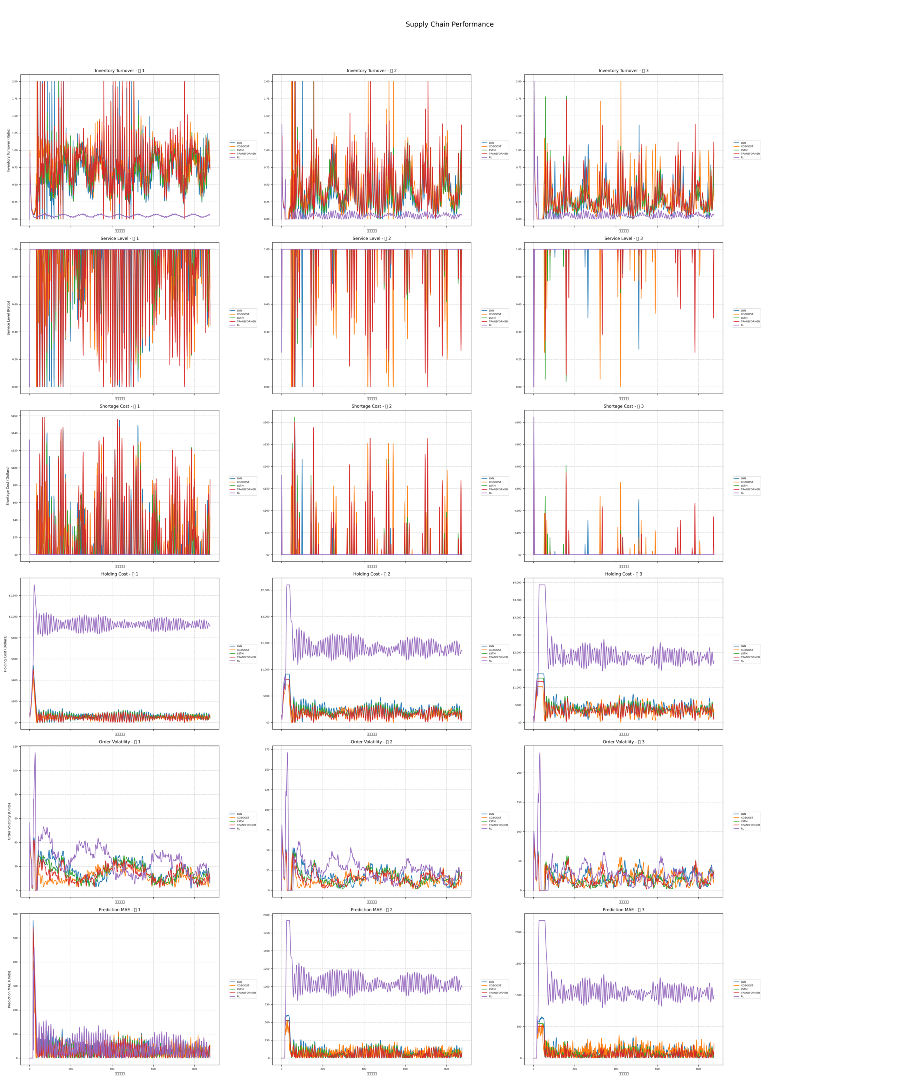}
    \caption{Combined Metrics for Supply Chain Performance}
    \label{fig:all-metrics}
\end{figure}

\begin{figure}[htbp]
    \centering
    \vspace*{-8mm}
    \includegraphics[width=0.85\linewidth]{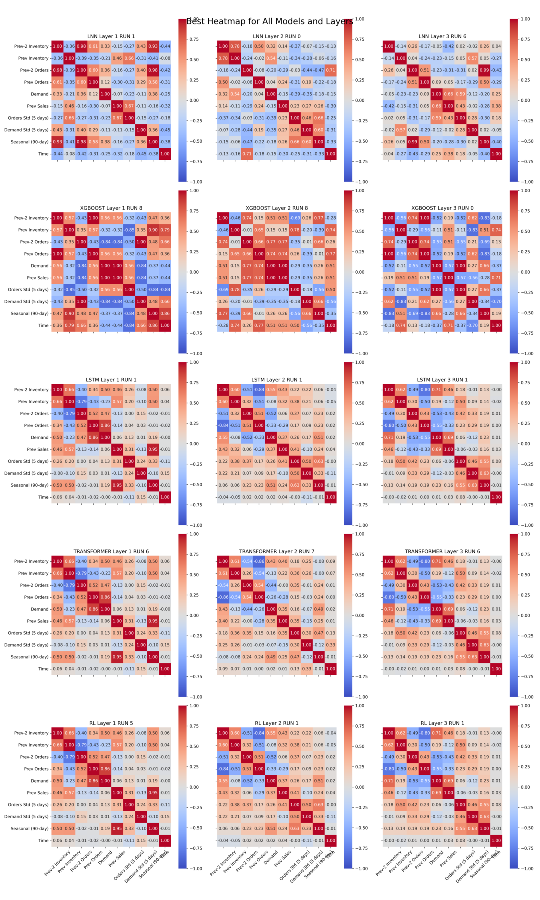}
    \caption{Heatmap of Model Performance Across All Layers}
    \label{fig:heatmap}
\end{figure}

The following figures provide additional visualizations of the supply chain performance metrics and model comparisons. Figure~\ref{fig:all-metrics} illustrates the combined metrics for supply chain performance across all models and layers, while Figure~\ref{fig:heatmap} presents a heatmap of model performance across all layers.

\end{document}